\documentclass[a4paper, 10pt, conference]{ieeeconf}      

\IEEEoverridecommandlockouts                              

\usepackage[T1]{fontenc}
\usepackage{cite}
\usepackage{amsmath,amssymb,amsfonts}
\usepackage{bbm}
\usepackage{graphicx}
\usepackage{subcaption}
\usepackage{tabularx}
\usepackage{booktabs}
\usepackage{multirow}
\usepackage{siunitx}
\usepackage{hyperref}
\hypersetup{
    colorlinks=true,
    linkcolor=blue,
    filecolor=magenta,      
    urlcolor=blue,
}
\usepackage{definitions}
\usepackage[absolute,overlay]{textpos}

\title{\LARGE \bf
Generalizing Decision Making for Automated Driving with an Invariant Environment Representation using Deep Reinforcement Learning}

\author{Karl Kurzer$^{1}$, Philip Sch\"orner$^{2}$, Alexander Albers$^{3}$, Hauke Thomsen$^{3}$, Karam Daaboul$^{1}$ and J. Marius Z\"ollner$^{1}$
\thanks{$^{1}$Karlsruhe Institute of Technology, Kaiserstr. 12, 76131 Karlsruhe, Germany
        {\tt\small \{kurzer, daaboul, marius.zoellner\}@kit.edu}}%
\thanks{$^{2}$FZI Research Center for Information Technology, Haid-und-Neu-Str. 10-14, 76131 Karlsruhe, Germany
        {\tt\small schoerner@fzi.de}}%
\thanks{$^{3}$These authors contributed equally {\tt\small \{alexander.albers, hauke.thomsen\}@student.kit.edu}}
}

\begin{document}
\begin{textblock*}{\textwidth}(19mm,10mm)
	\footnotesize
	\noindent \copyright 2021 IEEE.  Personal use of this material is permitted.  Permission from IEEE must be obtained for all other uses, in any current or future media, including reprinting/republishing this material for advertising or promotional purposes, creating new collective works, for resale or redistribution to servers or lists, or reuse of any copyrighted component of this work in other works.\\
	\textit{2021 IEEE Intelligent Vehicles Symposium (IV)}
\end{textblock*}
\maketitle

\thispagestyle{empty}
\pagestyle{empty}

\begin{abstract}
Data driven approaches for decision making applied to automated driving require appropriate generalization strategies, to ensure applicability to the world's variability. Current approaches either do not generalize well beyond the training data or are not capable to consider a variable number of traffic participants.
Therefore we propose an invariant environment representation from the perspective of the ego vehicle. The representation encodes all necessary information for safe decision making.
To assess the generalization capabilities of the novel environment representation, we train our agents on a small subset of scenarios and evaluate on the entire diverse set of scenarios. Here we show that the agents are capable to generalize successfully to unseen scenarios, due to the abstraction. In addition we present a simple occlusion model that enables our agents to navigate intersections with occlusions without a significant change in performance.
\end{abstract}

\section{INTRODUCTION}
Due to the increase in computational power, modern research for decision making and control has shifted from a model-based towards data-driven approach. While this paradigm shift has shown promising results, it requires careful consideration when it comes to the abstraction of the input data. This implies, that the designer of the input representation must not over-fit to the data, in order to generalize well to unseen data.

\begin{figure}
\centering
\includegraphics[width=\columnwidth]{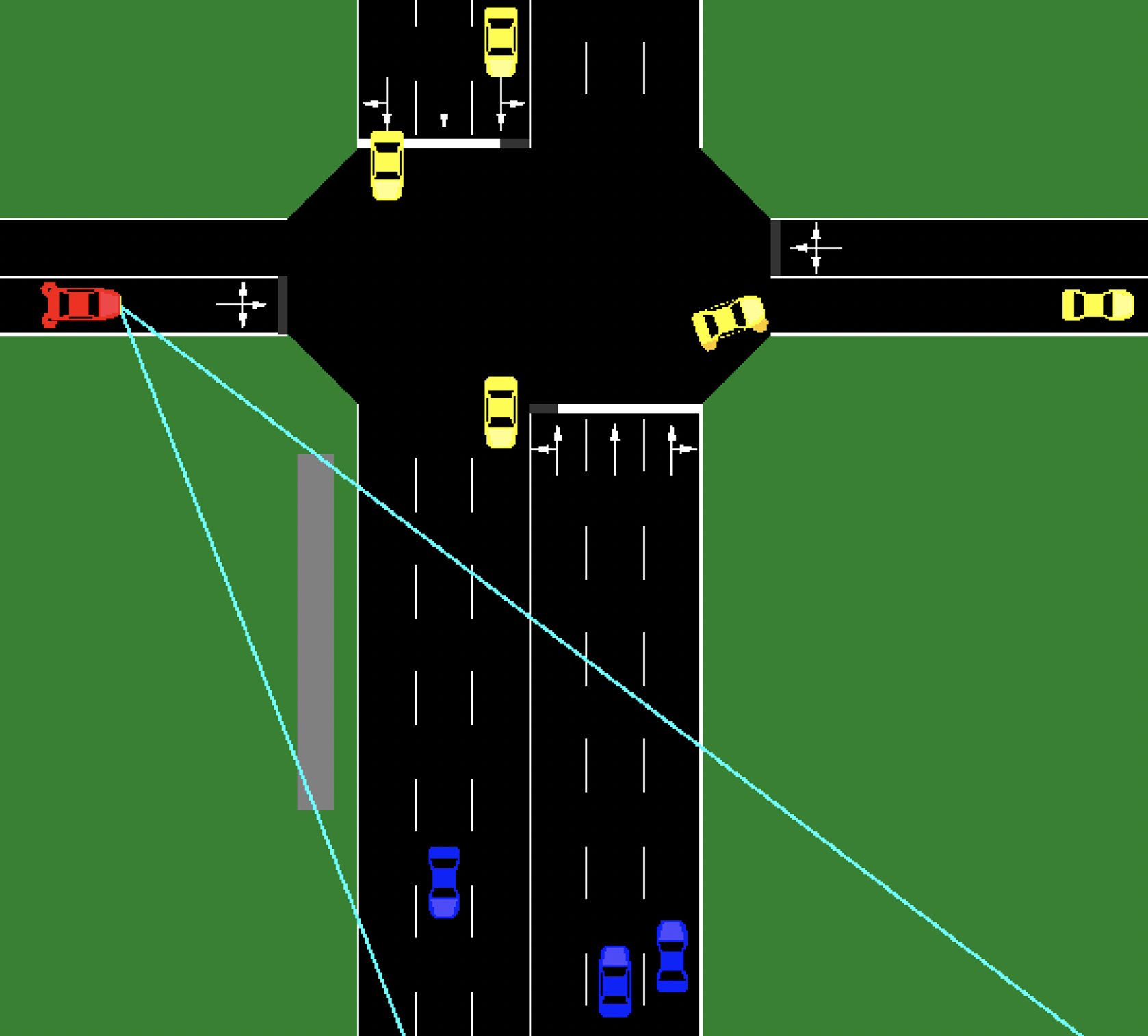}
\caption{Unsignalized intersection scenario with multiple lanes, perceivable traffic participants (yellow) as well as occluded traffic participants (blue) from the view of the ego vehicle (red), which needs to yield to other traffic; The ego vehicle needs to control its velocity in order to safely cross the intersection.}
\label{fig:main}
\end{figure}

Finding the right environment representation is especially important for areas such as automated driving. While the task of driving remains largely the same all over the world, the environment might look very different. One of the biggest challenges in this domain is the crossing of unsignalized intersections, cf. Fig.~\ref{fig:main}. Conventional model-based approaches are often used for solving simple driving problems, but quickly reach their limits in more complex driving environments, since all possible situations must be anticipated and implemented in advance. A common heuristic strategy is to use the Time-To-Collision (TTC) \cite{Hayward1972} as a threshold to determine when unsignalized intersections may be crossed safely. Despite its benefits of being reliable and safe for a large number of situations, it requires manual tuning and can be difficult to scale in environments with a large number of road users and incomplete information. Especially in urban areas the complexity rises quickly and intersections are often not completely observable due to occlusions.

Therefore, the problem of crossing intersections is frequently learned with the help of Deep Reinforcement Learning \cite{Isele2017a, Isele2018,Kamran2020, Kai2020, Bouton2019}, a reinforcement learning method that uses deep neural networks. In the past, Deep Reinforcement Learning has shown an immense potential in solving high-dimensional sequential decision making problems such as Atari \cite{Mnih2013,Mnih2015}, leading up to agents capable of reaching super-human performance on a variety of extremely challenging tasks \cite{Schrittwieser2019}.
However, learning based approaches often require fixed dimensions of the input data. Hence, in the context of automated driving, the input is often limited to a fixed number of traffic participants to be considered or to a fixed discretized bird's eye view of the environment that contains a large portion of irrelevant information.

In this work, we propose an invariant environment representation (IER), that is independent of the road layout and the number of traffic participants. Further, we train various agents using Deep Reinforcement Learning on a small set of intersection layouts and demonstrate their capability of generalizing the learned behavior to arbitrary intersection layouts.

\section{Related Work}
Safe decision making and thus correct behavior at intersections poses a major challenge for autonomous vehicles, especially when dealing with occlusions.
Recent approaches often apply Partially Observable Markov Decision Processes (POMDP) to account for  occluded areas.
Model-based approaches make assumptions about potentially occluded traffic participants \cite{Hubmann2019,Schorner2019}.
They demonstrate that POMDPs are suitable to deal with these situations.
However, the size of the state space depends on the number of the known and the assumed traffic participants.

Other approaches applied to intersection scenarios which are based on Reinforcement Learning, frequently use a fixed number of traffic participants \cite{Tram2018, Kamran2020, Kai2020}. In addition, some represent occlusions with phantom vehicles \cite{Kamran2020}, and others account for changing behaviors of other traffic participants over time with recurrent neural networks \cite{Tram2018}.

To obtain a fixed size of the state space, Isele et al. rasterize the environment using discretized Cartesian coordinates \cite{Isele2017a, Isele2018}.
For occupied cells, the cells contain information about the expected Time-To-Collision, the normalized heading angles as well as velocity of the corresponding vehicle.
They also infer the optimal policy using Reinforcement Learning and validate their approach in five perpendicular intersection scenarios with a varying number of crossing lanes.

Bouton et al. present a different approach to deal with multiple objects to be considered using a scalable decision making process based on reward decomposition \cite{Bouton2017}.
Two different offline POMDP solvers are evaluated in scenarios with occlusions.
The computation time grows linearly with the number of other traffic participants.
Another work employs scene decomposition to account for multiple traffic participants in their approach for safe reinforcement learning applying QMDPs \cite{Bouton2019}.
Uncertainties, such as occlusions or perception errors, are tracked with a learned belief tracker.
The output of the reinforcement algorithm is checked by a model checker to ensure safe outputs.
Due to the grid-based representation of the environment the latter approaches add information to the state space that does not contribute to the decision making process.

Although the existing approaches achieve promising results and suggest the applicability of Reinforcement Learning, the ability to allow generalization is still lacking, which in turn is crucial for data-driven methods. The current state of research either restricts the intersection layouts to similar right-angled intersections or limits the number of traffic participants that can be observed by the agent.
To address these limitations, we propose an invariant environment representation which is relative to the ego vehicle and accounts for an arbitrary number of vehicles.

\section{PROBLEM STATEMENT}
The problem of intersection navigation is formulated as a Markov Decision Process (MDP).
The ego vehicle, being the agent, chooses an action in each time step.
Then the agent collects an immediate reward and the system is transferred to the
next state.

The MDP is described by the tuple $\langle \statespace, \actionspace, \transitionmodel, \rewardmodel, \discountfactor\rangle$ \cite{Sutton2018}.
\begin{itemize}
	\item $\statespace$ denotes the \emph{state space} of the agent. 
	\item $\actionspace$ denotes the \emph{action space} of the agent. 
	\item $\transitionmodel: \statespace \times \actionspace \times \statespace \to [0,1]$ is the
	\emph{transition function} with $P(\state' | \state, \action)$ specifying the probability of a 
	transition from state \state{} to state $\state'$	given the action $\action$ is chosen by the agent.
	\item $\rewardmodel: \statespace \times \actionspace \times \statespace \to \mathbb{R}$ is the 
	reward function with $r(\state, \state', \action)$ denoting the resulting reward of the action 
	$\action$ taken in state $\state$.
	\item $\discountfactor \in [0,1]$ denotes a \emph{discount factor} controlling the influence of future rewards on the current state. 
\end{itemize}
The policy $\policy$ of the agent is a mapping from the state to the probability of each available action and is given by $\policy: \statespace\times 
\actionspace \to [0,1]$.

The goal of each agent is to maximize its expected cumulative reward in the MDP, starting from its current state:
$G = \sum {\gamma^t r(\state, \state', \action)}$ where $t$ denotes the time and $G$ the return, representing the cumulated discounted rewards.
$V(\state)$ is called the state-value function, given by $V^\policy(\state) = E[G|\state,\policy]$.
Similarly, the action-value function $Q(\state, \action)$ is defined as $Q^\policy(\state,\action) = E[G|\state,\action]$,
representing the expected return of choosing action $\action$ in state $\state$.

The optimal policy starting from state $s$ is defined as $\policy^* = \argmax_{\policy} V^\policy(\state)$. 
The state-value function is optimal under the optimal policy: $max\ V = V^{\policy^*}$. The same is 
true for the action-value function: $max \ Q=Q^{\policy^*}$.
The optimal policy can be discovered by maximizing over $Q^*(\state,\action)$:
\begin{equation}\label{Eq:OptPolicy}
\policy^{*}(a|s) = \left\{
\begin{array}{rcl}
1, & {\mbox{if}} \ {a = \argmax_{a \in {\cal A}} Q^{*}(\state,\action)}  \\
0, & \mbox{otherwise}
\end{array}\right..
\end{equation}
Once $Q^*$ has been determined, the optimal policies can easily be derived. Thus, it is the goal to learn the optimal action-value function $Q^*(\state,\action)$ for all possible state-action combinations of the MDP.

\section{APPROACH}
Learning this action-value function can be achieved with Q-Learning. We specifically make use of Deep Q-Learning, which uses function approximation via Deep Neural Networks to estimate the action-value function for all possible state-action combinations \cite{Mnih2013, Mnih2015}.

While automated driving as a whole is a hard task, the domain of intersections is particularly complex. During regular driving, a lane following behavior is adapted, where the focus is on the vehicle directly in front. At intersections however, traffic participants can cross the driving path from many directions at different distances simultaneously, hence planning needs to incorporate not just one but multiple traffic participants from different angles. As intersections offer a high degree of complexity, the timing of crossing needs to be precise to ensure safety.
In order to learn a policy that is invariant given the environment, i.e. it is generally applicable to any intersection layout, the input representation needs to discard information regarding the road layout.

Therefore, we propose an input representation relative to the ego vehicle and its trajectory using a Frenet coordinate system \cite{Werling2010}, in order to ensure an invariant representation of the environment.

\subsection{Invariant Environment Representation}\label{Sec:IER}
\begin{figure}
\centering
\includegraphics[width=\columnwidth]{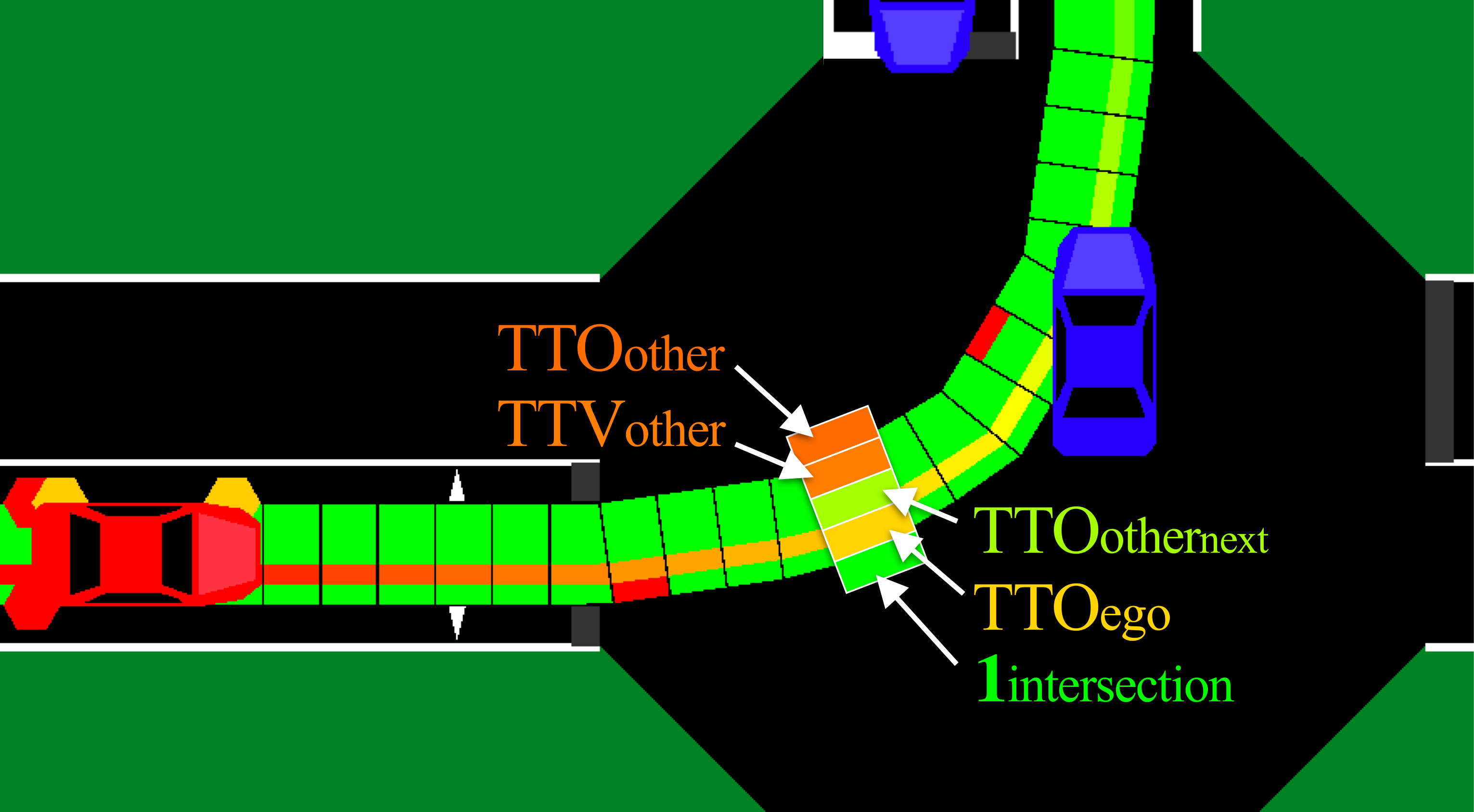}
\caption{Each patch encodes the time when it is going to be occupied by another vehicle, \tto{} and when it is vacant again, \ttv{}, as well as \tton{}
and \ttoe{} for the ego vehicle. In addition \ibit{} is used to indicate that an intersection starts at the respective patch (red). For illustrative purposes, the TTO and TTV sections in the patches are using a color scale from red ($t \to 0$) to green ($t \to 1$).}
\label{fig:IER}
\end{figure}

The invariant environment representation (IER) is inspired by other occupancy based approaches, which represent navigable space employing a map representation with specific values \cite{Isele2017a,Isele2018,Kurzer2020}. 
The path in front of the ego vehicle is discretized into so called patches. The discretization length depends on the desired look ahead distance.
However, it does not only encode the occupancy information, but includes the following values, namely the \Ttto{} (TTO) and the \Tttv{} (TTV), which are both based on the current velocity (constant velocity assumption\footnote{More advanced approaches for the velocity prediction of other traffic participants could easily be employed, but were not explored.}) of other traffic participants as well as the ego vehicle's velocity,  see Fig.~\ref{fig:IER}. The \Ttto{} (TTO) refers to the time\footnote{The term time is used as a synonym for duration.} when a patch becomes occupied, the \Tttv{} (TTV) to the time when a patch becomes vacant.

The TTO is defined by \eqref{Eq:TTO}:
\begin{equation}\label{Eq:TTO}
TTO = \frac{s_{\text{start}}}{v},
\end{equation}
where $s_\text{start}$ is the distance on the traffic participant's path until the first intersection of the traffic participant (i.e. the front) with the corresponding patch, and $v$ the velocity of the traffic participant.

The TTV is defined by \eqref{Eq:TTV}:
\begin{equation}\label{Eq:TTV}
TTV = \frac{s_\text{end}}{v},
\end{equation}
where $s_\text{end}$ is the distance on the traffic participant's path until the last intersection of the traffic participant (i.e. the rear) with the patch.

In order for the IER to be applicable for an arbitrary number of vehicles, multiple traffic participants are unified to one, if the time gap between two traffic participants traveling on the same path is less than a velocity dependent threshold, i.e. there is no space to pass safely. This implies that the \Tttv{} always belongs to the last traffic participant of the union.
In addition, we use the indicator function \ibit{} to encode information about the start of the intersection for some experiments, see \ref{sec:training}.

Each patch of the IER is fully described by the following values:
\begin{itemize}
	\item \tto\\\Ttto{} of this patch by other traffic
	\item \ttv\\\Tttv{} of this patch by other traffic
	\item \tton\\Next \Ttto{} of this patch by other traffic after the last traffic participant of the union
	\item \ttoe\\\Ttto{} of this patch by the ego vehicle
	\item \ibit\\Intersection bit indicating whether this patch is the first patch of an intersection
\end{itemize}
In this work, the IER partitions the driving path in front of the ego vehicle along its center line into 50 patches of a length of \SI{1}{\meter} and a width corresponding to the width of the ego vehicle. During our work we learned, that we achieved a higher success rate, when we encoded only the first patch of an intersection with another vehicle's driving path along the ego vehicle's driving path. Hence we chose this encoding for all experiments.
Both \Ttto{} and \Tttv{} are clamped and normalized with $t_{max} = \SI{10}{\second}$, cf. \eqref{Eq:Scaling}. The value of $t_{max}$ corresponds to a look-ahead of almost \SI{140}{\meter} at typical urban velocities of \SI{50}{\kilo\meter\per\hour}.
In case the patch of the ego vehicle's driving path does not intersect with another path, the respective values for the \Ttto{} and the \Tttv{} are set to $1$:
\begin{equation}\label{Eq:Scaling}
t = \frac{\min(t,t_{max})}{t_{max}} \in [0,1].
\end{equation}
We use a simple occlusion model, using a worst case assumption enabling the agent to navigate intersections with occluded vehicles. Occluded areas of the lanes are assumed to be occupied by a vehicle with the length of the occlusion on the respective lane. Further, we model this occluded vehicle driving at the speed limit. The encoding in the IER commences in the same way as for non occluded vehicles.

\subsection{Action Space}
As we employ Deep Q-Learning, we require a discrete action space. The agent can choose among the following actions in order to control its velocity.
Each action is being executed over a step length of \SI{0.4}{\second}.
\begin{itemize}
    \item Accelerate: \SI{3}{\meter\per\second\squared}
    \item Maintain: \SI{0}{\meter\per\second\squared}
    \item Decelerate: \SI{-3}{\meter\per\second\squared}
\end{itemize}

\subsection{Reward Function}
The reward function \eqref{Eq:Reward} consists of three distinct parts, namely a reward for collision, velocity and acceleration:
\begin{equation}\label{Eq:Reward}
r(\state, \state',\action) = r_\text{collision}(s') + r_\text{velocity}(s) + r_\text{acceleration}(a).
\end{equation}
The collision reward \eqref{Eq:RewardCollision} includes all collision states as well as states that are deemed a near-collision, i.e. the agent crosses the path of a traffic participant with less than \SI{10}{\meter} lateral clearance or is approaching a traffic participant in its path closer than \SI{1}{\meter} in longitudinal direction:
\begin{equation}\label{Eq:RewardCollision}
r_\text{collision} = -k_c\left(\ibco \cup \mathbbm{1}_\text{near-collision}\right).
\end{equation}
To encourage the agent to drive close to the speed limit of \SI{50}{\kilo\meter\per\hour}, larger deviations from it are penalized in the velocity reward \eqref{Eq:RewardVelocity} with a greater penalty for exceeding the speed limit ($k_{v_\text{upper}} > k_{v_\text{lower}}$):
\begin{equation}\label{Eq:RewardVelocity}
r_\text{velocity} =\begin{cases}
-k_{v_\text{upper}}\left|v-v_\text{upper}\right|, &\text{if}\ v>v_\text{upper}\\
-k_{v_\text{lower}}\left|v-v_\text{lower}\right|, &\text{if}\ v<v_\text{lower}.
\end{cases}
\end{equation}
Lastly, the acceleration reward \eqref{Eq:RewardAcceleration} penalizes acceleration to foster a smoother driving style:
\begin{equation}\label{Eq:RewardAcceleration}
r_\text{acceleration} = -k_{a}\left|\action\right|.
\end{equation}

\section{EXPERIMENTS}
The experiments are conducted using the open source microscopic traffic simulator SUMO\footnote{\href{https://www.eclipse.org/sumo/}{https://www.eclipse.org/sumo/}} \cite{SUMO2018}.
With the Traffic Control Interface from SUMO, all information which is required to compute the IER described in \ref{Sec:IER} is retrieved at every simulation step.
In order to ensure a wide variability of traffic situations, each intersecting lane uses a randomized vehicle flow.
We are starting the simulation with a constant traffic density by setting the probability of emitting a vehicle per second to \SI{10}{\percent} per lane. This leads to a density of approximately \SI{7}{vehicles\per\kilo\meter}  and lane. The traffic flow is reduced after \SI{30}{\second} to ensure that situations arise where the agent could pass the intersection safely. This is done by reducing the emitting probability to \SI{5}{\percent} (approx. \SI{4}{vehicles\per\kilo\meter}) per lane.
It is important to note that all non-ego vehicles controlled by SUMO are set to ignore the ego vehicle. This means that they do not brake to avoid collisions. This is necessary in order for the agent to avoid learning a policy where it would exploit the fact that other agents brake for it, rather than solving the scenario due to its own actions.

\subsection{Scenarios}
\begin{figure*}
     \centering
     \begin{subfigure}[b]{0.3\textwidth}
         \centering
         \includegraphics[width=\textwidth]{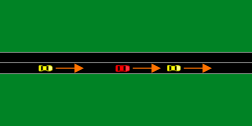}
         \caption*{(Sc01) Car Following}
         \label{fig:sc01}
     \end{subfigure}
     \hfill
     \begin{subfigure}[b]{0.3\textwidth}
         \centering
         \includegraphics[width=\textwidth]{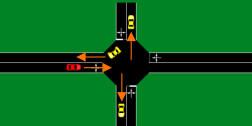}
         \caption*{\color{red}(Sc02) One Lane}
         \label{fig:sc02}
     \end{subfigure}
     \hfill
     \begin{subfigure}[b]{0.3\textwidth}
         \centering
         \includegraphics[width=\textwidth]{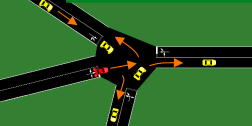}
         \caption*{(Sc03) One Lane Curved}
         \label{fig:sc03}
     \end{subfigure}
          \begin{subfigure}[b]{0.3\textwidth}
         \centering
         \includegraphics[width=\textwidth]{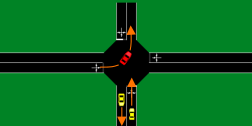}
         \caption*{(Sc04) Left Turn}
         \label{fig:sc04}
     \end{subfigure}
     \hfill
     \begin{subfigure}[b]{0.3\textwidth}
         \centering
         \includegraphics[width=\textwidth]{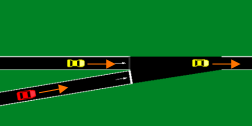}
         \caption*{(Sc05) Merge}
         \label{fig:sc05}
     \end{subfigure}
     \hfill
     \begin{subfigure}[b]{0.3\textwidth}
         \centering
         \includegraphics[width=\textwidth]{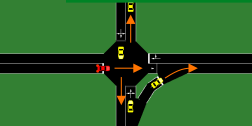}
         \caption*{(Sc06) Merge Intersection}
         \label{fig:sc06}
     \end{subfigure}
          \begin{subfigure}[b]{0.3\textwidth}
         \centering
         \includegraphics[width=\textwidth]{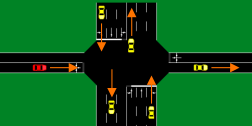}
         \caption*{\color{red}(Sc07) Three Lanes}
         \label{fig:sc07}
     \end{subfigure}
     \hfill
     \begin{subfigure}[b]{0.3\textwidth}
         \centering
         \includegraphics[width=\textwidth]{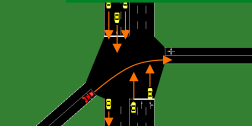}
         \caption*{(Sc08) Three Lanes Curved}
         \label{fig:sc08}
     \end{subfigure}
     \hfill
     \begin{subfigure}[b]{0.3\textwidth}
         \centering
         \includegraphics[width=\textwidth]{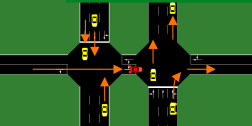}
         \caption*{(Sc09) Double Three Lanes}
         \label{fig:sc09}
     \end{subfigure}
        \caption{Scenarios used for {\color{red}training (red)} and evaluation. 
        (Sc01) Ego vehicle conducts car following;
        (Sc02) Ego vehicle crosses an intersection with one lane per direction; 
        (Sc03) Same as Sc02, but curved; 
        (Sc04) Ego vehicle turns left at an intersection with one lane per direction; 
        (Sc05) Ego vehicle merges into a continuous flow of vehicles; 
        (Sc06) Combination of Sc02 and Sc05; 
        (Sc07) Ego vehicle crosses an intersection with three lanes per direction; 
        (Sc08) Same as Sc07, but curved; 
        (Sc09) Same as Sc07, but with a subsequent intersection}
        \label{fig:scenarios}
\end{figure*}

Figure~\ref{fig:scenarios} depicts nine out of thirteen scenarios that the agents were evaluated on.
The scenarios were designed with the goal of creating a diverse set in order to be able to demonstrate the applicability of our IER to the varied situations that exist in the real world.
The complexity increases from scenario 1 to scenario 9. Scenario 10 and scenario 11 are variations of \hyperref[fig:sc01]{scenario 1} and \hyperref[fig:sc05]{scenario 5} with slower traffic. Scenario 12 and 13 are similar to \hyperref[fig:sc02]{scenario 2}, with scenario 12 being a Y-shaped one lane intersection with three directions and scenario 13 an intersection with two lanes from the top and one lane from the bottom.
In contrast to related work, we position our agent further away from the intersection (\SI{100}{\meter}).
This expands the task of the agent from a mere stop-or-go decision to velocity control, which results in a less jerky and more human-like decision making, considering how humans adapt their velocity when approaching an intersection with respect to the traffic. In addition, this expands the capabilities of our agent to car following scenarios, see Fig.~\ref{fig:scenarios}~\hyperref[fig:sc01]{(Sc01)}.

\subsection{Training} \label{sec:training}
With the goal to test the generalization capabilities of our IER, we train five different agents on a small subset (Sc02 and Sc07) of all thirteen scenarios.
The first three agents are trained on scenarios with increasing complexity and variability, but without occlusions and the \ibit{} missing in the state space.
Based on the results of these three agents, cf. Table~\ref{tab:generalization}, we choose the agent with the best combination of success rate and early termination rate for the training and evaluation with occlusions.
\begin{itemize}
    \item \ag{1}: Trained on an intersection with a single lane per direction (Fig.~\ref{fig:scenarios}~\hyperref[fig:sc02]{(Sc02)})
    \item \ag{2}: Trained on an intersection with three lanes per direction (Fig.~\ref{fig:scenarios}~\hyperref[fig:sc07]{(Sc07)})
    \item \ag{3}: Trained on scenario 2 and scenario 7
    \item \ag{4}: \ag{2} trained with occlusions, but without \ibit{}
    \item \ag{5}: \ag{2} trained without occlusions, but with \ibit{}
\end{itemize}
We utilize an $\epsilon\text{-greedy}$ policy and decay $\epsilon$ to $\epsilon_\text{final}$ over the course of the first \SI{30}{\percent} of the training. Additionally, we employ several powerful extensions to the original DQN \cite{Hessel18}, such as prioritized experience replay \cite{Schaul15} as well as Double-Q learning \cite{Hasselt16}, to improve the stability of the training and alleviate the over-estimation of action-values in Q-learning, respectively.

The most important hyperparameters for the DQN are summarized in Table~\ref{tab:hyperparameters}. The training is conducted using Stable Baselines \cite{stable-baselines} in combination with TensorFlow \cite{Tensorflow2015}.
\begin{table}
\centering
\caption{Hyperparameters of the Deep-Q Network}
\begin{tabularx}{\columnwidth}{XX}
\toprule
\textbf{Parameter}     & \textbf{Value} \\
\midrule
Learning rate         & 2e-4            \\
Discount factor       & 0.99            \\
$\epsilon_\text{final}$ & 0.05         \\
Buffer size           & 50,000          \\
Batch size            & 256             \\
Double-Q learning   & true            \\
Prioritized experience replay  & true            \\
MLP network size      & (60, 60)        \\
Training steps        & 5M           \\\bottomrule
\label{tab:hyperparameters}
\end{tabularx}
\end{table}

\begin{table}
\centering
\caption{Parameters used in the Environment}
\begin{tabularx}{\columnwidth}{XX}
\toprule
\textbf{Parameter}     & \textbf{Value} \\
\midrule
maximum steps per episode      & \SI{250}{}          \\
step length             & \SI{0.4}{\second}        \\
$v_\text{upper}$        & \SI[parse-numbers=false]{14.\overline{4}}{\meter\per\second}        \\
$v_\text{lower}$        & \SI[parse-numbers=false]{13.\overline{3}}{\meter\per\second}        \\
$k_{c}$                 & 115   \\
$k_{v_\text{upper}}$    & 0.03  \\
$k_{v_\text{lower}}$    & 0.01  \\ 
$k_{a}  $               & 0.002 \\
\bottomrule
\label{tab:environmentparameters}
\end{tabularx}
\end{table}

\section{EVALUATION}
All agents are evaluated on the scenarios presented in Fig.~\ref{fig:scenarios} as well as Sc10-Sc13. Each agent is evaluated for 1000 episodes on each scenario. Occlusions are positioned randomly on both the left and right side of the junction in \SI{100}{\percent} of all evaluation episodes, cf. Fig~\ref{fig:main}. A video with the performance of \ag{4} on Sc01-Sc09 can be found online \footnote{\href{https://url.kurzer.de/IER}{https://url.kurzer.de/IER}}. The parameters used for our environment are listed in Table~\ref{tab:environmentparameters}. For reference, we provide a tuned TTC-based agent, that performs similar to our best agent, cf. Table~\ref{tab:generalization} and Table~\ref{tab:occlusion}. Here, for each lane, the absolute difference between the \Ttto{} of the intersecting and the ego vehicle is calculated: $|\tto-\ttoe|$. The baseline agent accelerates to a maximum of \SI{50}{\kilo\meter\per\hour} whenever all differences are greater than \SI{1.6}{\second} and decelerates otherwise. In contrast to the constant velocity assumption of the RL agents, \ttoe{} is calculated for the TTC-based agent using a constant acceleration assumption.

\subsection{Metrics}
The key metric, which is used to evaluate the agent's performance, is the success rate (SR).
It describes the fraction of successful intersection traversals, i.e. reaching the other side of the intersection, while avoiding a collision or near-collision.
The frequency with which a collision or near-collision occurs is denoted as early termination rate (ETR).
Some agents adopt a policy to decelerate to a standstill before the intersection and wait for an opportunity to pass. When this happens the episode terminates after a maximum of 250 steps (with a step length of \SI{0.4}{\second}, resulting in a maximum episode duration of \SI{100}{\second}), thus: $1-\text{SR} \neq \text{ETR}$.

\subsection{Results}
The results are summarized in Table~\ref{tab:generalization} (generalization) and Table~\ref{tab:occlusion} (occlusion). Since other vehicles are not braking for the ego vehicle, some collisions would not occur in real traffic. While this results in a fairly high early termination rate, it does not affect the comparability of the trained agents.

With regard to the generalization capabilities it can be seen, that \ag{1}, which was trained solely on Sc02, performs worse than \ag{2} and \ag{3}. It outperforms \ag{2} solely on scenarios with a single lane per direction, which was to be expected.
The best performing agent is \ag{2} that was trained on Sc07 (when considering SR and ETR at the same time). This is likely due to the higher complexity in the experienced states. Similarly, \ag{3} performs well, and improves upon \ag{2} in single lane scenarios.
Overall, \ag{1} seems less cautious, since its early termination rate is considerably higher than it is for the other agents, probably because of the less challenging traffic situation depicted in \hyperref[fig:sc02]{(Sc02)}. Furthermore, this indicates why there is an increase of almost \SI{48}{\percent} in early terminations from \ag{2} to \ag{3}.

The results show that the agents performance on previously unseen scenarios is comparable to the performance of the training scenarios, given that the complexity in the training scenario is sufficiently high. This provides a strong indication that their ability to generalize from a small subset of scenarios to the others is due to the invariant environment representation.

The evaluation with occlusions demonstrates the usefulness of our simple occlusion model. The performance of \ag{2} drops considerably, which has not experienced any occlusions during training. In contrast, \ag{4} performs best overall, which was to be expected since it has experienced occlusions during training. The agent, which was trained using \ibit{}, \ag{5}, performs significantly better than the \ag{2}. Without occlusions its mean SR and ETR are \SI{81.1}{\percent} and \SI{13.4}{\percent} respectively, indicating that it is beneficial for the agent to know where the intersection starts in case of occlusions. From the sum of the SR and the ETR in Table~\ref{tab:occlusion}, we can further derive that \ag{2} and \ag{5} wait more often before the intersection for a gap, instead of creeping into the intersection, contrasting the results without occlusions, see Table~\ref{tab:generalization}. This leads to more episodes where the maximum number of steps is reached. The \ag{4} implicitly learns where to expect crossing vehicles and how to smartly explore the intersection. In contrast to the other agents, almost no episode ended due to reaching the end of the episode. During the exploration, the agent creeps into the intersection area in order to perceive the occluded lanes. It should be noted, that the state space of \ag{5} is larger. Hence a larger network might improve performance, however, this was not evaluated.
\begin{table}
\caption{Generalization: Success Rate (SR) and Early Termination Rate (ETR)}
\begin{tabularx}{\columnwidth}{X*{6}{r}}
\toprule
\textbf{Scenario}   & \textbf{Metric}  & \textbf{\ag{1}}  & \textbf{\ag{2}}  & \textbf{\ag{3}} & \textbf{\ag{TTC}}\\
\toprule
\multirow{2}{*}{Sc01}     & SR  & \SI{99.2}{\percent}   & \SI{91.3}{\percent}   & \SI{100.0}{\percent} & \SI{100.0}{\percent}  \\
                          & ETR  & \SI{0.2}{\percent}    & \SI{7.2}{\percent}    & \SI{0.0}{\percent}  & \SI{0.0}{\percent}   \\ \midrule
\multirow{2}{*}{Sc02}     & SR  & \SI{87.4}{\percent}   & \SI{85.3}{\percent}   & \SI{86.2}{\percent}  & \SI{100.0}{\percent}  \\
                          & ETR  & \SI{11.8}{\percent}   & \SI{9.5}{\percent}    & \SI{13.8}{\percent} & \SI{0.0}{\percent}   \\ \midrule
\multirow{2}{*}{Sc03}     & SR  & \SI{80.4}{\percent}   & \SI{83.7}{\percent}   & \SI{81.2}{\percent}  & \SI{96.0}{\percent}  \\
                          & ETR  & \SI{18.8}{\percent}   & \SI{10.9}{\percent}   & \SI{18.8}{\percent} & \SI{4.0}{\percent}   \\ \midrule
\multirow{2}{*}{Sc04}     & SR  & \SI{90.3}{\percent}   & \SI{89.8}{\percent}   & \SI{91.6}{\percent}  & \SI{98.5}{\percent}  \\
                          & ETR  & \SI{8.9}{\percent}    & \SI{5.3}{\percent}    & \SI{8.4}{\percent}  & \SI{1.5}{\percent}   \\ \midrule
\multirow{2}{*}{Sc05}     & SR  & \SI{92.7}{\percent}   & \SI{88.5}{\percent}   & \SI{95.1}{\percent}  & \SI{88.0}{\percent}  \\
                          & ETR  & \SI{6.5}{\percent}    & \SI{5.5}{\percent}    & \SI{4.9}{\percent}  & \SI{12.1}{\percent}   \\ \midrule
\multirow{2}{*}{Sc06}     & SR  & \SI{83.2}{\percent}   & \SI{81.3}{\percent}   & \SI{86.3}{\percent}  & \SI{87.9}{\percent}  \\
                          & ETR  & \SI{16.1}{\percent}   & \SI{13.7}{\percent}   & \SI{13.7}{\percent} & \SI{12.3}{\percent}   \\ \midrule
\multirow{2}{*}{Sc07}     & SR  & \SI{58.0}{\percent}   & \SI{83.9}{\percent}   & \SI{74.1}{\percent}  & \SI{73.0}{\percent}  \\
                          & ETR  & \SI{41.2}{\percent}   & \SI{10.9}{\percent}   & \SI{25.9}{\percent} & \SI{27.0}{\percent}   \\ \midrule
\multirow{2}{*}{Sc08}     & SR  & \SI{60.5}{\percent}   & \SI{75.3}{\percent}   & \SI{69.4}{\percent}  & \SI{78.5}{\percent}  \\
                          & ETR  & \SI{38.9}{\percent}   & \SI{16.2}{\percent}   & \SI{30.7}{\percent} & \SI{21.5}{\percent}   \\ \midrule
\multirow{2}{*}{Sc09}     & SR  & \SI{52.8}{\percent}   & \SI{64.8}{\percent}   & \SI{63.3}{\percent}  & \SI{85.7}{\percent}  \\
                          & ETR  & \SI{46.5}{\percent}   & \SI{20.7}{\percent}   & \SI{36.7}{\percent} & \SI{14.4}{\percent}   \\ \midrule
\multirow{2}{*}{Sc10}     & SR  & \SI{39.3}{\percent}   & \SI{97.2}{\percent}   & \SI{100.0}{\percent} & \SI{88.5}{\percent}  \\
                          & ETR  & \SI{60.7}{\percent}   & \SI{2.8}{\percent}    & \SI{0.0}{\percent}  & \SI{11.5}{\percent}   \\ \midrule
\multirow{2}{*}{Sc11}     & SR  & \SI{80.9}{\percent}   & \SI{80.4}{\percent}   & \SI{92.8}{\percent}  & \SI{90.4}{\percent}  \\
                          & ETR  & \SI{16.4}{\percent}   & \SI{5.8}{\percent}    & \SI{7.2}{\percent}  & \SI{9.7}{\percent}   \\ \midrule
\multirow{2}{*}{Sc12}     & SR  & \SI{89.8}{\percent}   & \SI{89.6}{\percent}   & \SI{93.8}{\percent}  & \SI{82.7}{\percent}  \\
                          & ETR  & \SI{9.4}{\percent}    & \SI{5.5}{\percent}    & \SI{6.2}{\percent}  & \SI{17.4}{\percent}   \\ \midrule
\multirow{2}{*}{Sc13}     & SR  & \SI{77.3}{\percent}   & \SI{83.7}{\percent}   & \SI{81.1}{\percent}  & \SI{71.0}{\percent}  \\
                          & ETR  & \SI{22.0}{\percent}   & \SI{11.0}{\percent}   & \SI{18.9}{\percent} & \SI{29.0}{\percent}   \\ \midrule
\multirow{2}{*}{Mean}     & SR  & \SI{76.3}{\percent}   & \SI{84.2}{\percent}   & \SI{85.8}{\percent}  & \SI{87.7}{\percent}  \\
                          & ETR  & \SI{22.9}{\percent}   & \SI{9.6}{\percent}    & \SI{14.2}{\percent} & \SI{12.3}{\percent}   \\ \bottomrule
\end{tabularx}
\label{tab:generalization}
\end{table}
\begin{table}
\caption{Occlusions: Success Rate (SR) and Early Termination Rate (ETR)}
\begin{tabularx}{\columnwidth}{X*{6}{r}}
\toprule
\textbf{Scenario}   & \textbf{Metric}  & \textbf{\ag{2}}  & \textbf{\ag{4}}  & \textbf{\ag{5}} & \textbf{\ag{TTC}}\\
\toprule
\multirow{2}{*}{Sc01}     & SR  & \SI{91.3}{\percent}   & \SI{100.0}{\percent}  & \SI{96.0}{\percent}  & \SI{100.0}{\percent} \\
                          & ETR  & \SI{7.2}{\percent}    & \SI{0.0}{\percent}    & \SI{0.0}{\percent}  & \SI{0.0}{\percent}  \\ \midrule
\multirow{2}{*}{Sc02}     & SR  & \SI{70.9}{\percent}   & \SI{91.0}{\percent}   & \SI{85.7}{\percent}  & \SI{100.0}{\percent} \\
                          & ETR  & \SI{7.5}{\percent}    & \SI{9.0}{\percent}    & \SI{10.4}{\percent} & \SI{0.0}{\percent}  \\ \midrule
\multirow{2}{*}{Sc03}     & SR  & \SI{60.7}{\percent}   & \SI{87.0}{\percent}   & \SI{62.0}{\percent}  & \SI{93.7}{\percent} \\
                          & ETR  & \SI{8.0}{\percent}    & \SI{12.3}{\percent}   & \SI{10.2}{\percent} & \SI{6.3}{\percent}  \\ \midrule
\multirow{2}{*}{Sc04}     & SR  & \SI{86.2}{\percent}   & \SI{93.0}{\percent}   & \SI{86.0}{\percent}  & \SI{98.3}{\percent} \\
                          & ETR  & \SI{3.9}{\percent}    & \SI{7.0}{\percent}    & \SI{10.0}{\percent} & \SI{1.7}{\percent}  \\ \midrule
\multirow{2}{*}{Sc05}     & SR  & \SI{88.2}{\percent}   & \SI{94.9}{\percent}   & \SI{91.1}{\percent}  & \SI{86.7}{\percent} \\
                          & ETR  & \SI{5.6}{\percent}    & \SI{5.1}{\percent}    & \SI{4.7}{\percent}  & \SI{13.0}{\percent}  \\ \midrule
\multirow{2}{*}{Sc06}     & SR  & \SI{73.9}{\percent}   & \SI{89.5}{\percent}   & \SI{79.2}{\percent}  & \SI{87.1}{\percent} \\
                          & ETR  & \SI{10.5}{\percent}   & \SI{10.6}{\percent}   & \SI{16.9}{\percent} & \SI{13.1}{\percent}  \\ \midrule
\multirow{2}{*}{Sc07}     & SR  & \SI{52.9}{\percent}   & \SI{77.3}{\percent}   & \SI{76.4}{\percent}  & \SI{65.2}{\percent} \\
                          & ETR  & \SI{7.4}{\percent}    & \SI{22.4}{\percent}   & \SI{17.1}{\percent} & \SI{24.2}{\percent}  \\ \midrule
\multirow{2}{*}{Sc08}     & SR  & \SI{63.7}{\percent}   & \SI{72.1}{\percent}   & \SI{47.1}{\percent}  & \SI{71.3}{\percent} \\
                          & ETR  & \SI{12.4}{\percent}   & \SI{27.9}{\percent}   & \SI{26.8}{\percent} & \SI{22.0}{\percent}  \\ \midrule
\multirow{2}{*}{Sc09}     & SR  & \SI{41.1}{\percent}   & \SI{63.3}{\percent}   & \SI{59.1}{\percent}  & \SI{77.0}{\percent} \\
                          & ETR  & \SI{13.4}{\percent}   & \SI{32.1}{\percent}   & \SI{34.7}{\percent} & \SI{13.0}{\percent}  \\ \midrule
\multirow{2}{*}{Sc10}     & SR  & \SI{97.2}{\percent}   & \SI{100.0}{\percent}  & \SI{99.1}{\percent}  & \SI{89.1}{\percent} \\
                          & ETR  & \SI{2.8}{\percent}    & \SI{0.0}{\percent}    & \SI{0.9}{\percent}  & \SI{10.8}{\percent}  \\ \midrule
\multirow{2}{*}{Sc11}     & SR  & \SI{80.3}{\percent}   & \SI{91.7}{\percent}   & \SI{80.4}{\percent}  & \SI{89.0}{\percent} \\
                          & ETR  & \SI{7.0}{\percent}    & \SI{8.3}{\percent}    & \SI{10.7}{\percent} & \SI{11.2}{\percent}  \\ \midrule
\multirow{2}{*}{Sc12}     & SR  & \SI{76.6}{\percent}   & \SI{94.0}{\percent}   & \SI{87.6}{\percent}  & \SI{79.2}{\percent} \\
                          & ETR  & \SI{4.2}{\percent}    & \SI{6.0}{\percent}    & \SI{8.4}{\percent}  & \SI{16.6}{\percent}  \\ \midrule
\multirow{2}{*}{Sc13}     & SR  & \SI{49.6}{\percent}   & \SI{83.1}{\percent}   & \SI{80.0}{\percent}  & \SI{56.6}{\percent} \\
                          & ETR  & \SI{6.2}{\percent}    & \SI{14.6}{\percent}   & \SI{15.8}{\percent} & \SI{25.0}{\percent}  \\ \midrule
\multirow{2}{*}{Mean}     & SR  & \SI{71.7}{\percent}   & \SI{87.5}{\percent}   & \SI{79.2}{\percent}  & \SI{84.1}{\percent} \\
                          & ETR  & \SI{7.4}{\percent}    & \SI{11.9}{\percent}   & \SI{12.8}{\percent} & \SI{12.1}{\percent}  \\ \bottomrule
\end{tabularx}
\label{tab:occlusion}
\end{table}

\section{CONCLUSION}
In this work we presented a novel input representation for an autonomous agent to learn the task of intersection navigation.
The abstraction of the environment through our invariant environment representation enables our trained agents to transfer their knowledge gathered on one intersection to arbitrary intersection layouts. This is especially useful considering the variety of layouts an autonomous agent can experience in the real world. Further, using a simple occlusion model our representation is applicable to intersections that are not fully observable.

The promising results have spurred us to conduct further research on the topic.
Currently we focus on a continuous action space using Soft-Actor-Critic methods \cite{Haarnoja2018} as well as the encoding of adaptive behavior \cite{Wolf2018}.
In addition, the current input representation is being extended.

\section*{ACKNOWLEDGMENT}
We wish to thank the German Research Foundation (DFG) for funding the project  Cooperatively  Interacting  Automobiles  (CoInCar)  within  which  the research leading to this contribution was conducted. The information as well as  views  presented  in  this  publication  are  solely  the  ones  expressed  by  the authors.

\addtolength{\textheight}{-2.0cm}   

\def\url#1{}
\bibliographystyle{IEEEtran}
\bibliography{library}

\end{document}